\documentclass[11pt,a4paper]{article}
\usepackage[hyperref]{emnlp2020}
\usepackage{times}
\usepackage{latexsym}

\usepackage{graphicx}
\usepackage{subcaption}
\usepackage{amssymb}
\usepackage{amsmath}
\usepackage[utf8]{inputenc}
\usepackage[english]{babel}
\usepackage{verbatim}

\usepackage{microtype}
\DeclareMathOperator{\EX}{\mathbb{E}}

\DeclareMathAlphabet\mathbfcal{OMS}{cmsy}{b}{n}

\aclfinalcopy 
\def\aclpaperid{425} 

\title{Control, Generate, Augment: \\A Scalable Framework for Multi-Attribute Text Generation}

\author{Giuseppe Russo\textsuperscript{1}, Nora Hollenstein\textsuperscript{1}, Claudiu Musat\textsuperscript{2}, Ce Zhang\textsuperscript{1} \\
\textsuperscript{1} ETH Zurich, \{russog,noraho,ce.zhang\}@inf.ethz.ch \\
\textsuperscript{2} Swisscom, claudiu.musat@swisscom.com \\}

\date{}

\begin{document}
\maketitle
\begin{abstract}

We introduce CGA, a conditional VAE architecture, to \textbf{c}ontrol, \textbf{g}enerate, and \textbf{a}ugment text. CGA is able to generate natural English sentences controlling multiple semantic and syntactic attributes by combining adversarial learning with a context-aware loss and a cyclical word dropout routine. We demonstrate the value of the individual model components in an ablation study. The scalability of our approach is ensured through a single discriminator, independently of the number of attributes. We show high quality, diversity and attribute control in the generated sentences through a series of automatic and human assessments. As the main application of our work, we test the potential of this new NLG model in a data augmentation scenario. In a downstream NLP task, the sentences generated by our CGA model show significant improvements over a strong baseline, and a classification performance often comparable to adding same amount of additional real data. 
\end{abstract}

\section{Introduction}


Recently, natural language generation (NLG) has become a prominent research topic in NLP due to its diverse applications, ranging from machine translation (e.g., \newcite{sennrich2016controlling}) to dialogue systems (e.g., \newcite{budzianowski2019hello}). The common goal of these applications using automatic text generation is the augmentation of datasets used for supervised NLP tasks. To this end, one of the key demands of NLG is \textit{controlled} text generation, more specifically, the ability to systematically control semantic and syntactic aspects of generated text.

Most previous approaches simplify this problem by approximating NLG with the control of one single attribute of the text, such as sentiment or formality (e.g., \newcite{li2018delete}, \newcite{fu2018style}, and \newcite{john2019disentangled}). However, the problem of controlled generation typically relies on multiple components such as lexical, syntactic, semantic and stylistic aspects. Therefore, the simultaneous control of multiple attributes becomes vital to generate natural sentences suitable for downstream tasks. Methods such as the ones presented by \newcite{hu2017toward} and \newcite{subramanian2018multiple} succeed in simultaneously controlling various attributes. However, these methods depend on the transformation of input reference sentences, or do not scale easily to more than two attributes due to architectural complexities, such as the requirement for separate discriminators for each additional attribute.

In light of these challenges, we propose the \textbf{C}ontrol, \textbf{G}enerate, \textbf{A}ugment framework (CGA), a powerful model to synthesize additional labeled data sampled from a latent space. The accurate multi-attribute control of our approach offers significant performance gains on downstream NLP tasks. We provide the code and all generated English sentences to facilitate future research\footnote{\url{https://github.com/DS3Lab/control-generate-augment}}.

The main contribution of this paper is a scalable model which learns to control \textbf{multiple semantic and syntactic attributes} of a sentence. The CGA model requires only a single discriminator for simultaneously controlling multiple attributes. To the best of our knowledge, we are the first to incorporate techniques such as cyclical word-dropout and a context-aware loss, which allow the CGA model to generate natural sentences given a latent representation and an attribute vector, \textit{without requiring an input reference sentence during training}. We present automatic and human assessments to confirm the multi-attribute control and high quality of the generated sentences. Further, we provide a thorough comparison to previous work.

\begin{figure*}[ht!]
\small
\center
\includegraphics[width=0.85\textwidth]{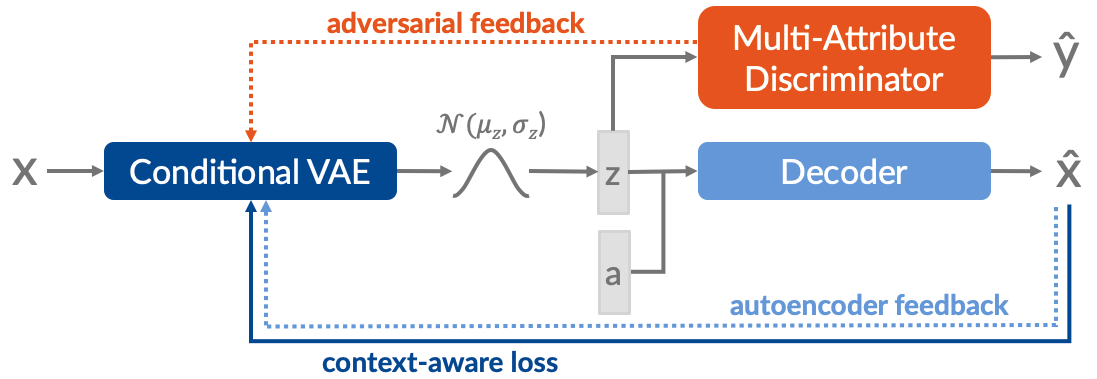}
\caption{Model architecture depicting the key components of CGA.}\label{fig:arch}
\end{figure*}
    
We use CGA as a natural language generation method for \textbf{data augmentation}, which boosts the performance of downstream tasks. We present data augmentation experiments on various English datasets, where we significantly outperform a strong baseline and achieve a performance often 
comparable to adding same amount of additional real data.

\section{Method}\label{sec:method}

We now present our model for controlled text generation.
Our model is based on the Sentence-VAE framework by \newcite{bowman2016generating}. However, we modify this model to allow the generation of sentences conditioned not only on the latent code but also on an attribute vector.
We achieve this by disentangling the latent code from the attribute vector, in a similar way as the Fader networks~\cite{lample2017fader}, originally developed for computer vision tasks. As we will see, this simple adaption is not sufficient, and we introduce further techniques to improve the multi-attribute sentence generation.

\subsection{Model Architecture}

We assume access to a corpus of sentences $\mathcal{X}=\{x_i\}_{i=1}^{N}$ and a set of $K$ categorical attributes of interest. For each 
sentence $x_i$, we use an attribute vector $a_i$ to represent these $K$
associated attributes.
Example attributes include 
the sentiment or verb tense of a sentence.

Given a latent representation $z$, which encodes the context information of the corpus and an attribute vector $a$, our goal is to construct a ML model which generates a new sentence $x$ containing the attributes of $a$.

\paragraph{Sentence Variational Autoencoder} 

The main component of our model is  a Variational Auto-Encoder \cite{kingma2013auto}. The encoder network $E_{\theta_{enc}}$,
parameterized by a trainable parameter $\theta_{enc}$,
takes as input a sentence $x$
and defines a probabilistic distribution over the latent code $z$:
\begin{equation}
\small
\begin{aligned}
    z \thicksim E_{\theta_{enc}}(x) := q_{E}(z|x; \theta_{enc})
\label{eq:eq1}
\end{aligned}
\end{equation}

\noindent The decoder $G_{\theta_{dec}}$,
parameterized by a trainable parameter $\theta_{dec}$,
tries to reconstruct the input sentence $x$ from a latent code $z$ and its attribute vector $a$.
We assume that the 
reconstructed sentence $\hat{x}$ has the same number of 
tokens as the input sentence $x$:

\begin{equation}
\small
\begin{aligned}
    \hat{x} \thicksim G_{\theta_{dec}}(z,a) := p_{G}(\hat{x}|z,a; \theta_{dec})\\  =\prod_{t=1}^{T}
    p_{G}(\hat{x}_{t+1}|\hat{x}_{t}, z, a;\theta_{dec})
\label{eq:eq2}
\end{aligned}
\end{equation}
where $T$ is the length of the input sentence and $\hat{x}_{t}$
is the $t^{th}$ token.
Here we 
use $p_G$ to denote both 
sentence-level probability and
word-level conditional probability.

To train the encoder and
decoder, we use the following 
VAE loss: 
\begin{equation}
\small
\begin{aligned}
    L_{VAE}(\theta_{enc} ,\theta_{dec}) :=  KL(q_E(z|x)||p(z)) -\\ \EX_{z \thicksim q_E(z|x)}\log p_{G}(x|z,a; \theta_{dec}),
\label{eq:eq3}
\end{aligned}
\end{equation}
where $p(z)$ is a standard Gaussian distribution.

When we try to optimize the loss in Equation~\ref{eq:eq3}, the KL term often vanishes. This problem is
known in text generation
as \textit{posterior collapse} \cite{bowman2016generating}.
To mitigate this problem we 
follow \newcite{bowman2016generating}
and add a weight $\lambda_{kl}$ to the KL term in Equation~\ref{eq:eq3}. At the start of training, we set the weight to zero, so that the model learns to encode as much information in $z$ as possible. Then, as training progresses, we gradually increase this weight,
as in the standard KL-annealing
technique.

Moreover, the posterior collapse
problem occurs partially due to the fact that,
during training, our decoder $G_{\theta_{dec}}$ predicts each token conditioned on the previous \textit{ground-truth} token. We aim to make the model rely more on $z$. A natural way to achieve this is to weaken the decoder by removing some or all of this conditional information during the training process. Previous 
works~\cite{bowman2016generating, hu2017toward} replace a --- randomly selected --- significant portion of the ground-truth tokens with \texttt{UNKNOWN} tokens.
However, this can severely affect the decoder and deteriorate the generative capacity of the model.
Therefore, we define a new 
word-dropout routine, which aims at both accommodating 
the posterior collapse
problem and preserving the decoder capacity. Instead of fixing the word-dropout rate to a large constant value as in \newcite{bowman2016generating}, we use a cyclical word-dropout rate $\zeta$:
\begin{equation}
\small
\begin{aligned}
        \zeta(s) =
        \left\{ \begin{array}{ll}
            k_{max} & s\leq \text{warm-up} \\ 
            k_{max} & \big| \cos(\frac{2\pi}{\tau}s)\big| \geq k_{max}\\ 
            k_{min}  & \big| \cos(\frac{2\pi}{\tau}s)\big| \leq k_{min}\\ 
            \big|cos(\frac{2\pi}{\tau}s)\big| & \text{otherwise}
        \end{array} \right.
\label{eq:wd}
\end{aligned}
\end{equation}
where $s$ is the current training iteration, $k_{max}$ and $k_{min}$ are fixed constant values we define as upper and lower thresholds, and $\tau$ defines the period of the cyclical word-dropout rate schedule (see Suppl. Section \ref{suppl:training}).

\paragraph{Disentangling Latent Code $z$ and Attribute Vector $a$}
To be able to generate sentences given any attribute
vector $a'$, we have to disentangle the attribute vector with the latent code. In other words, we seek that $z$ is 
{\em attribute-invariant}:
A latent code $z$ is 
{\em attribute-invariant}
if given two sentences $x_1$ and
$x_2$, they only differ in their attributes (e.g., two  versions of the same review expressing opposite sentiment). Hence, they should result in the same latent representation $z$ = $E_{\theta_{enc}}(x_1)$ = $E_{\theta_{enc}}(x_2)$.

To achieve this, we use a concept from predictability minimization \cite{schmidhuber1992learning} and  adversarial training for domain adaptation
\cite{ganin2016domain,louppe2017learning}, which was recently applied in the Fader Networks by \newcite{lample2017fader}.
We apply adversarial learning directly on the latent code $z$ of the input sentence $x$. We set a min-max game and introduce 
a discriminator 
$D_{\theta_{disc}}(z)$, that takes as input the
latent code and tries to 
predict the attribute vector $a$.
Specifically, 
$D_{\theta_{disc}}(z)$ outputs
for each attribute $k$,
a probability distribution
$p_D^k$ over all its 
possible values.
To train the discriminator, we optimize for the following loss:
\begin{equation}
\small
\begin{aligned}
L_{DISC}(\theta_{disc})
:= -\log \prod_{k} p_D^k(a_k)
\label{eq:4}
\end{aligned}
\end{equation}
where $a_k$ is the ground-truth of the $k^{th}$ attribute.

Simultaneously, we hope to learn an encoder and decoder which  
 (1) combined with the attribute vector $a$, allows the decoder to reconstruct the input sentence $x$, and (2) does not allow the discriminator to infer the correct attribute vector corresponding to $x$. 
 We optimize for:
\begin{equation}
\small
\begin{aligned}
     L_{ADV} := L_{VAE}(\theta_{enc} ,\theta_{dec}) -
      \lambda_{Disc} L_{DISC}(\theta_{disc})
\label{eq:5}
\end{aligned}
\end{equation}

\paragraph{Context-Aware Loss}
Equation \ref{eq:5} forces our model to choose which information the latent code $z$ should retain or disregard. However, this approach comes with the risk of deteriorating the quality of the latent code itself. Therefore, inspired by \newcite{sanakoyeu2018style},
we propose an attribute-aware context loss, which tries to preserve the context information by comparing the sentence latent representation and its back-context representation:

\begin{equation}
\begin{aligned}
    L_{CTX} := \left \| E_{\theta_{enc}}(x)  -   E_{\theta_{enc}}(G_{\theta_{dec}}(E_{\theta_{enc}}(x)))\right \|_{1}
\end{aligned}\label{eq:ctx}
\end{equation}
We use a ``stop-gradient" procedure, i.e., we compute the gradient w.r.t. $E_{\theta_{enc}}(x)$, which makes the function in Equation \ref{eq:ctx} differentiable. 

The latent vector $z=E_{\theta_{enc}}(x)$ can be seen as a contextual representation of the input sentence $x$. This latent representation is changing during the training process and hence adapts to the attribute vector. Thus, when measuring the similarity between $z$ and the back-context representation $E_{\theta_{enc}}(G_{\theta_{dec}}(E_{\theta_{enc}}(x)))$, we focus on preserving those aspects which are profoundly relevant for the context representation. 

Finally, when training the 
encoder and decoder (given the current discriminator),
we optimize for the following loss:
\begin{equation}
\begin{aligned}
L_{CGA} := 
     L_{VAE}(\theta_{enc} ,\theta_{dec})
     + 
     \lambda_{CTX} L_{CTX} \\
     -
      \lambda_{Disc} L_{DISC}(\theta_{disc})
\end{aligned}
\end{equation}


\begin{table*}[t]
\centering
\begin{tabular}{ll}
\textbf{Sentence} & \textbf{Attributes} \\\hline
it was a great time to get the best in town and i loved it. & Past / Positive \\
it was a great time to get the food and it was delicious. & Past / Positive \\
it is a must! & Present/Positive \\
they're very reasonable and they are very friendly and helpful. & Present / Positive \\
i had a groupon and the service was horrible. & Past / Negative \\
this place was the worst experience i've ever had. & Past / Negative \\
it is not worth the money. & Present / Negative \\
there is no excuse to choose this place. & Present / Negative \\\hline
\end{tabular}
\caption{Examples of generated sentences with two attributes: \textsc{sentiment} and \textsc{verb tense}.}
\label{tab:tensesent}
\end{table*}

\begin{table*}[t!]
\centering
\begin{tabular}{ll}
\textbf{Sentence} & \textbf{Attributes} \\\hline
they have a great selection of beers and shakes. & Present / Positive / Plural \\
i love this place and i will continue to go here. & Present / Positive / Singular \\
the mashed potatoes were all delicious! & Past / Positive / Plural \\
the lady who answered was very friendly and helpful. & Past / Positive / Singular \\
the people are clueless. & Present / Negative / Plural \\
i mean i'm disappointed. & Present / Negative / Singular \\
drinks were cold and not very good. & Past / Negative / Plural \\
it was a complete disaster. & Past / Negative / Singular \\\hline
\end{tabular}
\caption{Examples of generated sentences with three attributes: \textsc{sentiment}, \textsc{verb tense}, and \textsc{person number}.}
\label{tab:tsn}
\end{table*}

\section{Evaluation}\label{sec:eval}

To assess our newly proposed model for controlled sentence generation, we perform the following evaluations described in this section: An automatic and human evaluation to analyze the quality of the new sentences with multiple controlled attributes; an examination of sentence embedding similarity to assess the diversity of the generated samples; downstream classification experiments with data augmentation on two different datasets to prove the effectiveness of the new sentences in a pertinent application scenario; and, finally, a comparison of our results to previous work to specifically contrast our model against other single and multi-attribute models.

\paragraph{Datasets} 
We conduct all experiments on two datasets, YELP and IMDB reviews. Both contain sentiment labels for the reviews. From the YELP business reviews dataset \cite{yelp2014data}, we use reviews only from the category \textit{restaurants}, which results in a dataset of approx. 600'000 sentences. The IMDB movie reviews dataset \cite{imdb2011} contains approx. 150'000 sentences. For reproducibility purposes, details about training splits and vocabulary sizes can be found in the supplementary materials (\ref{suppl:data}).

\begin{table}[t!]
\small
\centering
\begin{tabular}{|lccc|}
\hline
              & \textbf{Sentiment} & \textbf{Tense} & \textbf{Person} \\ \hline \hline
\textbf{YELP} & 91.1\% (0.04)             & 96.6\% (0.03)         & 95.9\% (0.06)                 \\
\textbf{IMDB} & 90.0\% (0.06)             &    96.1\% (0.04)      &         92.0\% (0.05)   \\\hline                  
\end{tabular}
\caption{Mean attribute matching accuracy of the 30K generated sentences (in \%); standard deviation reported in brackets.}
\label{tab:automatic_sentiment}
\end{table}

\paragraph{Attributes}
For our experiments we use three attributes: sentiment as a semantic attribute; verb tense and person number as syntactic attributes.\\

\noindent \textsc{Sentiment}: We labeled each review as \textit{positive} or \textit{negative} following \newcite{shen2017style}.

\noindent \textsc{Verb Tense}: We detect \textit{past} and \textit{present} verb tenses using SpaCy's part-of-speech tagging model\footnote{\url{https://spacy.io/usage/linguistic-features\#pos-tagging}}. We define a sentence as \textit{present} if it contains more present than past verbs. We provide the specific PoS tags used for the labeling in the supplementary materials (\ref{suppl:attr}).

\noindent \textsc{Person Number}: We also use SpaCy to detect \textit{singular} or \textit{plural} pronouns and nouns. Consequently, we label a sentence as \textit{singular} if it contains more singular than plural pronouns or nouns, we define it \textit{plural}, in the opposite case, \textit{balanced} otherwise. \\

\noindent We train our model to generate sentences of maximally 20 tokens by controlling one, two or three attributes simultaneously. The sentences are generated by the decoder as described in Equation \ref{eq:eq2}. We chose to set the maximum sentence length to 20 tokens in this work, since (a) it is considerably more than previous approaches (e.g., \newcite{hu2017toward} presented a max length of 15 tokens), and (b) it covers more than the 99th percentile of the sentence lengths in the datasets used, which is 16.4 tokens per sentence for YELP and 14.0 for IMDB.
In Table \ref{tab:tensesent} we illustrate some examples of sentences where the model controls two attributes, \textsc{Sentiment} and \textsc{Verb Tense}. Table \ref{tab:tsn} shows sentences where the model controls three attributes simultaneously. Sentences with single controlled attributes can be found in the supplementary material (\ref{suppl:sents}).

\paragraph{Experimental Setting}

The encoder and decoder are single-layer GRUs with a hidden dimension of 256 and maximum sample length of 20. The discriminator is a single-layer LSTM. To avoid a vanishingly small KL term in the VAE \cite{bowman2016generating}, we use a KL term weight annealing that increases from 0 to 1 during training according to a logistic scheduling. $\lambda_{disc}$ increases linearly from 0 to 20. Finally, we set the back-translation weight $\delta$ to 0.5.
All hyper-parameters are provided in the supplementary material (\ref{suppl:training}).

{}

\subsection{Quality of Generated Sentences}

\begin{table}[t!]
\centering
\begin{tabular}{|lrc|}
\hline
\textbf{Attribute} & \textbf{Sentences} & \textbf{Accuracy ($\kappa$)}  \\ \hline \hline
Sentiment          & 106 / 120            & 0.88 (0.73)                         \\
Verb Tense         & 117 / 120            & 0.98 (0.97)                  \\
Person Number      & 114 / 120            & 0.95 (0.85)                      \\ \hline
2 Attributes       & 120 / 120            & 1.0                               \\ \hline
3 Attributes       & 97 / 120             & 0.80                        \\ \hline
Coherence          & 79 / 120             & 0.66                 \\ \hline
\end{tabular}
\caption{Results of the human evaluation showing accuracy and Cohen's
$\kappa$ for each attribute. }
\label{tab:human-eval}
\vspace{-0.5cm}
\end{table}

We quantitatively measure the sentence attribute control of our CGA model by inspecting the accuracy of generating sentences containing the designated attributes by conducting both automatic and human evaluations.
\paragraph{Attribute Matching} For this automatic evaluation, we generate sentences given the attribute vector $a$ as described in Section \ref{sec:method}. To assign \textsc{Sentiment} attribute labels to the newly generated sentences, we apply a pre-trained TextCNN \cite{kim2014convolutional}.
To assign the \textsc{verb tense} and \textsc{person number} labels we use  SpaCy's part-of-speech tagging. 
We calculate the \textit{attribute matching accuracy} as the percentage of the predictions of these pre-trained models on the generated sentences that match the attribute labels expected to be generated by our CGA model. Table \ref{tab:automatic_sentiment} shows the averaged results over five balanced sets of 6000 sentences generated by CGA models, trained on YELP and IMDB, respectively.

\paragraph{Human Evaluation} To further understand the quality of the generated sentences we go beyond the automatic attribute evaluation and perform a human judgement analysis. We provide all generated sentences including the human judgements\footnote{\url{https://github.com/DS3Lab/control-generate-augment}}. One of our main contributions is the generation of sentences with up to three controlled attributes. Therefore, we randomly select 120 sentences generated by the CGA model trained on YELP, which controls all three attributes. Two human annotators labeled these sentences by marking which of the attributes are included correctly in the sentence. 

In addition to the accuracy we report inter-annotator rates with Cohen's $\kappa$. In 80\% of the sentences all three attributes are included correctly and in 100\% of the sentences at least two of the three attributes are present. Finally, the annotators also judged whether the sentences are grammatically correct, complete and coherent English sentences. Most of the incorrect sentences contain repeated words or incomplete endings. The results are shown in Table \ref{tab:human-eval}.

\begin{table*}[t!]
\centering 
\begin{tabular}{|l|ccc|ccc|}
\hline
                       & \multicolumn{3}{c|}{\textbf{YELP}}                       & \multicolumn{3}{c|}{\textbf{IMDB}}                           \\ \hline \hline
\textbf{Model}         & \textbf{Sentiment} & \textbf{Tense} & \textbf{Person} & \textbf{Sentiment} & \textbf{Tense} & \textbf{Person} \\ \hline
$L_{ADV}$ + \textit{standard WD }        & 88.5\%             & 95.5\%         & 92.7\%             & 84.1\%             & \textbf{97.0\%}         & 86.3\%                 \\
$L_{ADV}$ + \textit{cyclical WD}       & 89.7\%      &       \textbf{ 96.8\%}         & 93.1\%             & 88.0\%             & 93.6\%         & 91.4\%                 \\
$L_{CTX}$ + \textit{standard WD}             & 90.5\%             & 95.6\%         & 94.6\%             & 89.2\%             & 95.8\%         & 87.9\%                 \\
$L_{CTX}$ + \textit{cyclical WD} (\textbf{CGA})  & \textbf{91.1\%}             & 96.6\%         & \textbf{95.4\%  }           & \textbf{90.0\%    }         & 96.1\%         & \textbf{92.0\%     }            \\ \hline
\end{tabular}
\caption{Ablation Study of the key components, reporting attribute matching scores for three features on the YELP and IMDB datasets. \textit{$L_{ADV}$ + standard WD} is trained with the word-dropout of \newcite{bowman2016generating}; \textit{$L_{ADV}$ + cyclical WD} is trained with our cyclical word-dropout; \textit{$L_{CTX}$ + standard WD} is trained with the standard word-dropout and with context-aware loss; \textit{$L_{CTX}$ + cyclical WD} is trained with \textit{both} cyclical word-dropout and context-aware loss.}
\label{tab:ablation}
\end{table*}

\begin{figure}[h!]
\centering 
\begin{subfigure}{0.45\textwidth}
  \includegraphics[trim={0.5cm 0.5cm 0.5cm 1cm}, clip,width=\linewidth]{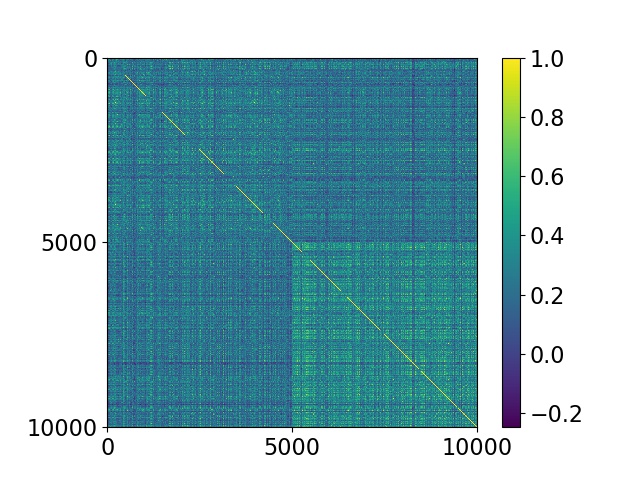}
  \subcaption{Real Data}
  \label{fig:sim1}
\end{subfigure}\hfil 
\begin{subfigure}{0.45\textwidth}
  \includegraphics[trim={0.5cm 0.5cm 0.5cm 1cm}, clip,width=\linewidth]{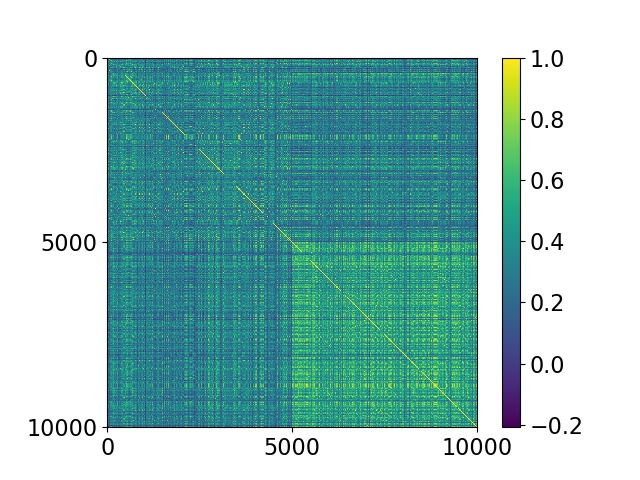}
  \subcaption{Generated Data}
  \label{fig:sim3}
\end{subfigure}
\caption{Similarity matrices for real data ($M_{real}$) and data generated by our CGA model controlling the sentiment attribute ($M_{gen}$).}\label{fig:similarity_matr}
\end{figure}

\begin{figure*}[ht!]
    \centering 
\begin{subfigure}{0.33\textwidth}
  \includegraphics[width=\linewidth]{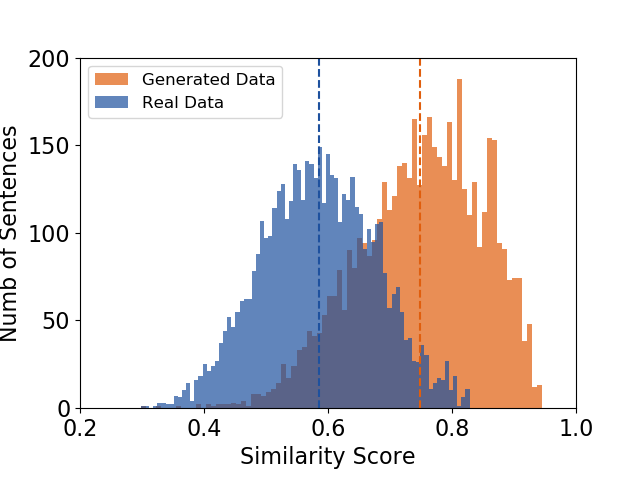}
  \caption{Negative-Negative}
  \label{fig:sim1}
\end{subfigure}\hfil 
\begin{subfigure}{0.33\textwidth}
  \includegraphics[width=\linewidth]{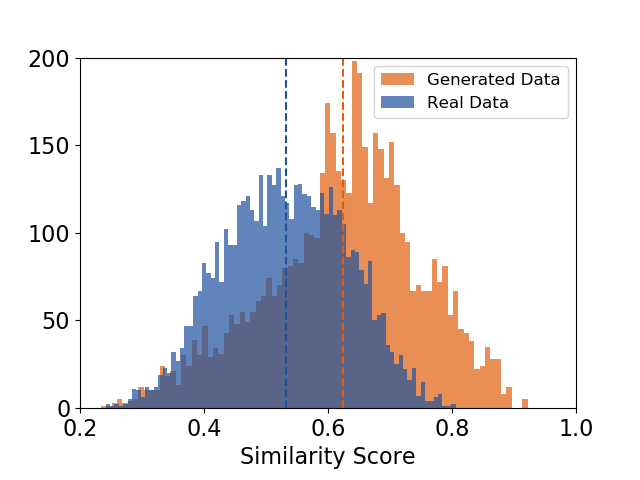}
  \caption{Negative-Positive}
  \label{fig:sim2}
\end{subfigure}\hfil 
\begin{subfigure}{0.33\textwidth}
  \includegraphics[width=\linewidth]{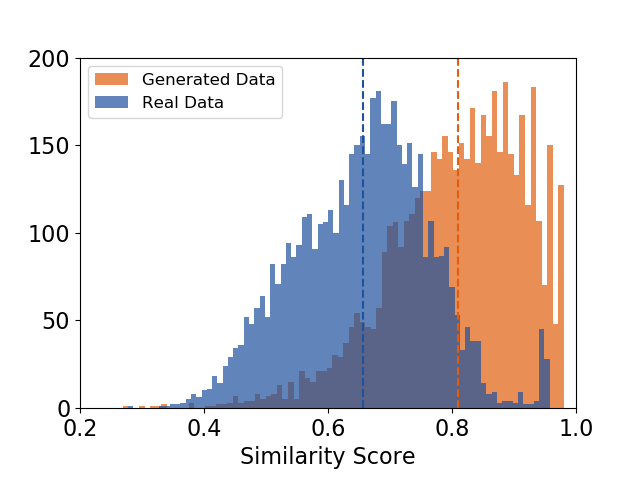}
  \caption{Positive-Positive}
  \label{fig:sim3}
\end{subfigure}
\caption{Sentence similarity scores computed for real data and data generated by our CGA model on the three sentiment clusters (Negative-Negative, Negative-Positive, Positive-Positive).}\label{fig:sim-hist}
\end{figure*}

\paragraph{Ablation Study}
We conduct an ablation study testing the key components of the CGA model, both on the YELP and IMDB datasets.
We separately trained four different versions of CGA to assess the impact on the multi-attribute control of the disjoint and joint usage of the \textit{cyclical word-dropout} (Equation \ref{eq:wd}) and of the \textit{context-aware loss} (Equation \ref{eq:ctx}). We computed the attribute matching score following the same approach described above. As shown in Table \ref{tab:ablation}, both techniques are beneficial for attribute control, especially for \textsc{sentiment} and \textsc{person number}. When the model is trained using at least one of these techniques it already shows significant improvements in all cases expect for \textsc{verb tense} on the IMDB data. 
Moreover, when the cyclical word-dropout and the context-aware loss are used jointly during training, the model experiences an increase of performance between 1-6\% w.r.t. the model trained without using these techniques.

\paragraph{Sentence Embedding Similarity}
Although generative models have been shown to produce outstanding results, in many circumstances they risk producing extremely repetitive examples (e.g., \newcite{zhao2017infovae}).
In this experiment, we qualitatively assess the capacity of our model to generate diversified sentences to further strengthen the results obtained in this work.
We sample 10K sentences from YELP ($D_{real}$) and from our generated sentences ($D_{gen}$), respectively, both labeled with the \textsc{sentiment} attribute. 
We retrieve the sentence embedding for each of the sentences in $D_{real}$ and $D_{gen}$ using the Universal Sentence Encoder \cite{cer2018universal}. Then, we compute the cosine similarity between the embeddings of all sentences of $D_{real}$ and, analogously, between the embeddings of our generated sentences $D_{gen}$.

Consequently, we obtain two similarity matrices $M_{real}$ and $M_{gen}$ (see Figure \ref{fig:similarity_matr}).
Both matrices show a four cluster structure: \textbf{top-left} -- similarity scores between negative reviews ($\mathcal{C}_{nn}$); \textbf{top-right or bottom-left} -- similarity scores between negative and positive reviews ($\mathcal{C}_{np}$); and \textbf{bottom-right} -- similarity scores between positive reviews ($\mathcal{C}_{pp}$).

Further, for each sample of $D_{real}$ and $D_{gen}$ we compute a similarity score as follows:

\begin{equation}
\begin{aligned}
    sim(s_{i,c}) = \frac{1}{K}\sum_{x \in \mathcal{N}_{K,c}}score(s_i, x)
\label{eq:simscore}
\end{aligned}
\end{equation}

where $c \in \{\mathcal{C}_{nn},\mathcal{C}_{np}, \mathcal{C}_{pp}\}$. $s_{i,}$ is the i-th sample of $D_{real}$ or $D_{gen}$ and $c$ is the cluster to which $s_{i}$ belongs. $\mathcal{N}_{K, c}$ is the set of the k-most similar neighbours of $s_{i}$ in cluster $c$, and $k$=50.

\begin{figure*}[t]
    \centering
    \includegraphics[width=\linewidth]{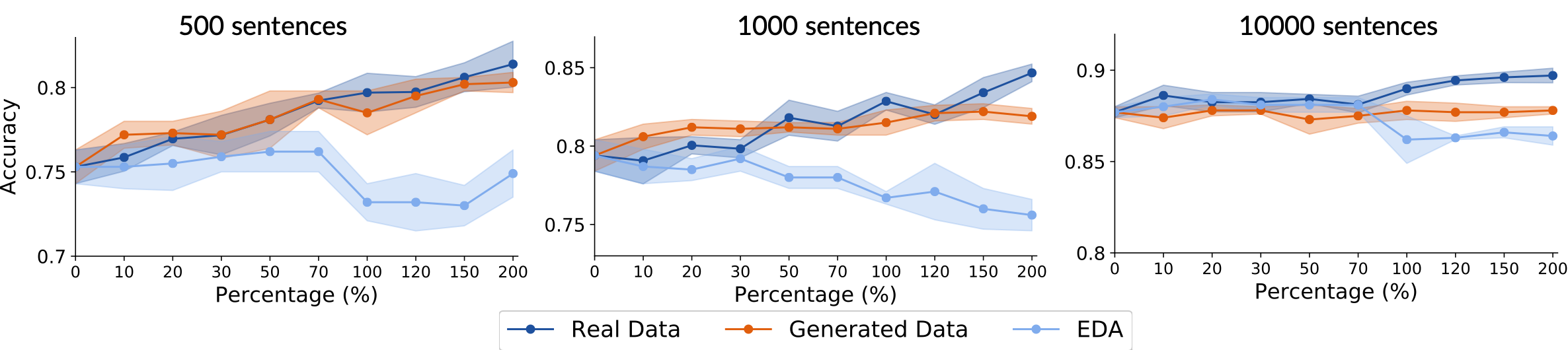}
    \caption{Data augmentation results for the YELP dataset.}
    \label{fig:dataug-yelp-line}
\end{figure*}

\begin{figure*}[t]
    \centering
    \includegraphics[width=\linewidth]{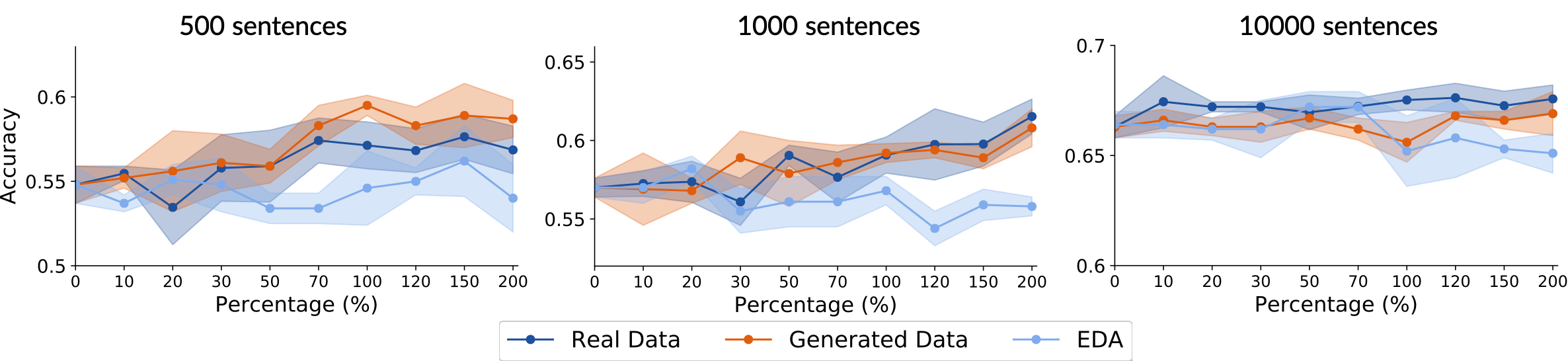}
    \caption{Data augmentation results for the IMDB dataset.}
    \label{fig:dataug-imdb-line}
\end{figure*}

\noindent To gain a qualitative understanding of the generation capacities of our model, we assume that an ideal generative model should produce samples that have comparable similarity scores to the ones or the real data. Figure \ref{fig:sim-hist} contrasts the similarity scores of $D_{real}$ and $D_{gen}$, computed on each cluster separately.

Although our generated sentences are clearly more similar between themselves than to the original ones, our model is able to produce samples clustered according to their labels. This highlights the good attribute control abilities of our CGA model and shows that it is able to generate diverse sentences which robustly mimic the structure of the original dataset. Hence, the generated sentences are good candidates for augmenting existing datasets.

We generalized this experiment for the multi-attribute case. The similarity matrices and the histograms for these additional experiments are provided in the supplementary material (\ref{suppl:embed}).

\begin{table*}[ht]
\centering
\begin{tabular}{|lccccclc|}
\hline
\multicolumn{1}{|c|}{} & \multicolumn{7}{c|}{\textbf{Training Size}} \\
 & \multicolumn{2}{|c|}{\textbf{500 sentences}} & \multicolumn{2}{c}{\textbf{1000 sentences}} & \multicolumn{3}{|c|}{\textbf{10000 sentences}} \\ \hline\hline
\multicolumn{1}{|l|}{\textbf{Model}} & \multicolumn{1}{c|}{acc. (std)} & \multicolumn{1}{c|}{\%} & \multicolumn{1}{c|}{acc. (std)} & \multicolumn{1}{c|}{\%} & \multicolumn{2}{c|}{10000} & \multicolumn{1}{c|}{\%} \\ \hline
\multicolumn{1}{|l|}{Real Data YELP} & \multicolumn{1}{c|}{0.75 (0.01)} & \multicolumn{1}{c|}{0} & \multicolumn{1}{c|}{0.79 (0.01)} & \multicolumn{1}{c|}{0} & \multicolumn{2}{c|}{0.87 (0.03)} & \multicolumn{1}{c|}{0} \\
\multicolumn{1}{|l|}{YELP + EDA} & \multicolumn{1}{c|}{0.77 (0.02)} & \multicolumn{1}{c|}{70} & \multicolumn{1}{c|}{0.80 (0.08)} & \multicolumn{1}{c|}{30} & \multicolumn{2}{c|}{0.88 (0.02)} & \multicolumn{1}{c|}{70} \\
\multicolumn{1}{|l|}{YELP + CGA (Ours)} & \multicolumn{1}{c|}{\textbf{0.80 (0.02)}} & \multicolumn{1}{c|}{150} & \multicolumn{1}{c|}{\textbf{0.82 (0.03)}} & \multicolumn{1}{c|}{120} & \multicolumn{2}{c|}{\textbf{0.88 (0.04)}} & \multicolumn{1}{c|}{100} \\ \hline\hline
\multicolumn{1}{|l|}{Real Data IMDB} & \multicolumn{1}{c|}{0.54 (0.01)} & \multicolumn{1}{c|}{0} & \multicolumn{1}{c|}{0.57 (0.06)} & \multicolumn{1}{c|}{0} & \multicolumn{2}{c|}{0.66 (0.05)} & \multicolumn{1}{c|}{0} \\
\multicolumn{1}{|l|}{IMDB + EDA} & \multicolumn{1}{c|}{0.56 (0.02)} & \multicolumn{1}{c|}{150} & \multicolumn{1}{c|}{0.58 (0.07)} & \multicolumn{1}{c|}{70} & \multicolumn{2}{c|}{0.67 (0.02)} & \multicolumn{1}{c|}{100} \\
\multicolumn{1}{|l|}{IMDB + CGA (Ours)} & \multicolumn{1}{c|}{\textbf{0.60 (0.01)}} & \multicolumn{1}{c|}{120} & \multicolumn{1}{c|}{\textbf{0.61 (0.01)}} & \multicolumn{1}{c|}{200} & \multicolumn{2}{c|}{\textbf{0.67 (0.03)}} & \multicolumn{1}{c|}{120} \\ \hline
\end{tabular}
\caption{Best performance for each method independent of the augmentation percentage used. For each method we report accuracy, standard deviation, and augmentation percentage.}
\label{tab:dataug-best}
\end{table*}

\subsection{Data Augmentation}\label{sec:dataaug}

The main application of our work is to generate sentences for data augmentation purposes. Simultaneously, the data augmentation experiments presented in this section reinforce the high quality of the sentences generated by our model.

As described, we conduct all experiments on two datasets, YELP and IMDB reviews. We train an LSTM sentiment classifier on both datasets, each with three different training set sizes. We run all experiments for training sets of 500, 1000 and 10000 sentences. These training sets are then augmented with different percentages of generated sentences (10, 20, 30, 50, 70, 100, 120, 150 and 200\%). This allows us to analyze the effect of data augmentation on varying original training set sizes, as well as varying increments of additionally generated data. In all experiments we average the results over 5 random seeds and we report the corresponding standard deviation.

To evaluate how beneficial our generated sentences are for the performance of downstream tasks, we compare training sets augmented with sentences generated from our CGA model to (a) real sentences from the original datasets, and (b) sentences generated with the Easy Data Augmentation (EDA) method by \newcite{wei2019eda}. EDA applies a transformation (e.g., synonym replacement or random deletion) to a given sentence of the training set and provides a strong baseline.

The results are presented in Figures \ref{fig:dataug-yelp-line} and \ref{fig:dataug-imdb-line}, for YELP and IMDB, respectively. They show the performance of the classifiers augmented with sentences from our CGA model, from EDA and from the original datasets. Our augmentation method proved to be favorable in all six scenarios. Our model clearly outperforms EDA in all the possible scenarios, especially with larger augmentation percentages. The performance of the classifiers augmented with CGA sentences is equal to real data, and only begins to diverge when augmenting the training set with more than 100\% of generated data.

In Table \ref{tab:dataug-best}, we report the best average test accuracy as well as the percentage of data increment of real data, EDA and our CGA model for all three training set sizes and both datasets. Numerical results for all augmentation percentages including validation performance can be found in the supplementary materials (\ref{suppl:results}).

\section{Comparison to Previous Work}\label{sec:comp}

As a final analysis, we compare our results with previous state-of-the-art models for both single-attribute and multi-attribute control.


\subsection{Single-Attribute Control}
\newcite{li2018delete} model style control in the \textit{Delete, Retrieve, Generate} framework, which deletes words related to a specific attribute and then inserts new words which belongs to the vocabulary of the target style (e.g., sentiment). \newcite{sudhakar2019transforming} improve this framework by combining it with a transformer architecture \cite{vaswani2017attention}. 
However, these approaches are susceptible to error, due to the difficulty of accurately selecting only the style-containing words.

Other approaches on text generation have leveraged adversarial learning. \newcite{john2019disentangled} use a VAE with multi-task loss to learn a content and style representation that allows to elegantly control the sentiment of the generated sentences while preserving the content. \newcite{shen2017style} train a cross-alignment auto-encoder (CAAE) with shared content and separate style distribution. \newcite{fu2018style} suggested a multi-head decoder to generate sentences with different styles. As in our work, \newcite{shen2017style} and \newcite{fu2018style} do not enforce content preservation. 

These models are specifically designed to control this single attribute by approximating the style of a sentence with its sentiment. \newcite{shen2017style} reported a sentiment matching accuracy of 83.5\%. Both \newcite{fu2018style} and \newcite{john2019disentangled} achieve better sentiment matching accuracy (96\% and 93.4\%, respectively) in the automatic evaluation than our CGA model trained for a single attribute (93.1\%). However, our CGA model obtains 96.3\% in human evaluation, which is comparable with these works. Moreover, CGA offers a strong competitive advantage because it guarantees high sentiment matching accuracy while controlling additional attributes and, thus, offers major control over multiple stylistic aspects of a sentence.

\subsection{Multi-Attribute Control}


Few works have succeed in designing an adequate model for text generation and controlling multiple attributes. \newcite{hu2017toward} use a VAE with controllable attributes to generate short sentences with max. 15 tokens. Our CGA model improves upon this by generating sentences of high quality with a max. length of 20 tokens. This restricted sentence length is still one of the major limitations in NLG.

\newcite{subramanian2018multiple} and \newcite{logeswaran2018content} apply a back-translation technique from unsupervised machine translation for style transfer tasks. \newcite{lai2019multiple} follow the approach of the CAAE with a two-phase training procedure. Unlike our CGA model, these works enforce content preservation and require input reference sentences. Hence, it is not straight-forward to directly compare the results. However, their reported attribute matching accuracies for the \textsc{sentiment} and \textsc{verb tense} attributes are considerably lower than ours (91.1\% and 96.6\%, respectively). CGA also yields significantly better performance in the human evaluation. Recently, \newcite{wang2019controllable} proposed an architecture for multi-attribute control. However, they focus merely on sentiment aspect attributes, while our CGA model is able to control both semantic \textit{and} syntactic attributes.

These previous works reported content preservation as an additional evaluation metric. It is important to note that this metric is of no interest for our work, since, differently from these previous models, CGA generates sentences directly from an arbitrary hidden representations and it does not need a reference input sentence. Moreover, our CGA model is scalable to more attributes, while the previous architectures require multiple discriminators for controlling the attributes. Although we provide extensive evaluation analyses, it is still an open research question to define an appropriate evaluation metric for text generation to allow for neutral comparisons.



\section{Conclusion}

To the best of our knowledge, we propose the first framework for controlled natural language generation which (1) generates coherent sentences sampling from a smooth latent space, with multiple semantic and syntactic attributes; (2) works within a lean and scalable architecture, and (3) improves downstream tasks by synthesizing additional labeled data.

To sum up, our CGA model, which combines a context-aware loss function with a cyclical word-dropout routine, achieves state-of-the-art results with improved accuracy on sentiment, verb tense and person number attributes in automatic and human evaluations. Moreover, our experiments show that our CGA model can be used effectively as a data augmentation framework to boost the performance of downstream classifiers.

A thorough investigation of the quality of the attribute-invariant representation in terms of independence between the context and the attribute vector will provide further insights. Additionally, a benchmark study of the maximum possible length of the generated sentences and the number of controllable attributes will deepen our understanding of the capabilities and limitations of CGA.

\section*{Acknowledgements}
CZ and the DS3Lab gratefully acknowledge the support from the Swiss National Science Foundation (Project Number 200021\_184628), Swiss Data Science Center, Alibaba, Cisco, eBay, Google Focused Research Awards, Oracle Labs, Swisscom, Zurich Insurance, Chinese Scholarship Council, and the Department of Computer Science at ETH Zurich.

\bibliography{emnlp2020}
\bibliographystyle{acl_natbib}

\newpage
\appendix

\begin{table*}[ht!]
\centering
\begin{tabular}{|l|cc|cc|ccc|}
\hline
                 & \multicolumn{2}{c|}{\textbf{Sentiment}} & \multicolumn{2}{c|}{\textbf{Tense}} & \multicolumn{3}{c|}{\textbf{Person Number}}                                 \\ \hline \hline
\textbf{Dataset} & \textbf{Positive}  & \textbf{Negative}  & \textbf{Present}   & \textbf{Past}  & \textbf{Singular} & \textbf{Plural} & \multicolumn{1}{l|}{\textbf{Balanced}} \\ \hline
\textbf{Yelp}    & 393,237             & 241,017             & 304,441             & 329,813         & 190,276            & 317,127          & 126,851                                \\
\textbf{IMDB}    & 71,676              & 64,625              & 86,965              & 69,336          & 53,468             & 54,046           & 28,787                                 \\ \hline
\end{tabular}
\caption{YELP and IMDB dataset details showing the exact number of sentences used and the class distribution.}
\label{tab:data}
\end{table*}

\section{Supplementary Material}

\subsection{Data}

\subsubsection{Structure}\label{suppl:data}
We use YELP and IMDB for the training, validation and testing of our CGA models. The label distributions for all attributes are described in Table \ref{tab:data}.

From the YELP business reviews dataset \cite{yelp2014data}\footnote{Retrieved from \url{https://github.com/shentianxiao/language-style-transfer}}, we use reviews only from the category \textit{restaurants}. We use the same splits for training, validation and testing as \newcite{john2019disentangled}, which contain 444101, 63483 and 126670, respectively. The vocabulary contains 9304 words.

We further evaluate our models on the IMDB dataset of movie reviews \cite{imdb2011}\footnote{Retrieved from \url{https://www.kaggle.com/lakshmi25npathi/imdb-dataset-of-50k-movie-reviews}}. We use reviews with less than 20 sentences and we select only sentences with less than 20 tokens. Our final dataset contains 122345, 12732, 21224 sentences for train validation and test, respectively. The vocabulary size is 15362 words.

\subsubsection{Attribute Labeling}\label{suppl:attr}
In this work we simultaneously control three attributes: \textsc{sentiment}, \textsc{verb tense} and \textsc{person number}.

We use SpaCy's Part-of-Speech tagging to assign the \textsc{verb tense} labels. Specifically, we use the tags \textsc{VBP} and \textsc{VBZ} to identify present verbs, and the tag \textsc{VBD} to identify past verbs.

Analogously, we use the SpaCy's PoS tags and the personal pronouns to assign \textsc{person number} labels.
In particular, we use the tag \textsc{NN}, which identifies singular nouns, and the following list of pronouns \{\textit{i, \textit{he}, \textit{she}, \textit{it}, \textit{myself}\}} to identify a singular sentence.
We use \textbf{NNS} and the list of pronouns \{\textit{we}, \textit{they}, \textit{themselves}, \textit{ourselves}\} to identify a plural sentence.

\begin{figure*}[ht!]
\centering 
\begin{subfigure}{0.5\textwidth}
  \includegraphics[width=\linewidth]{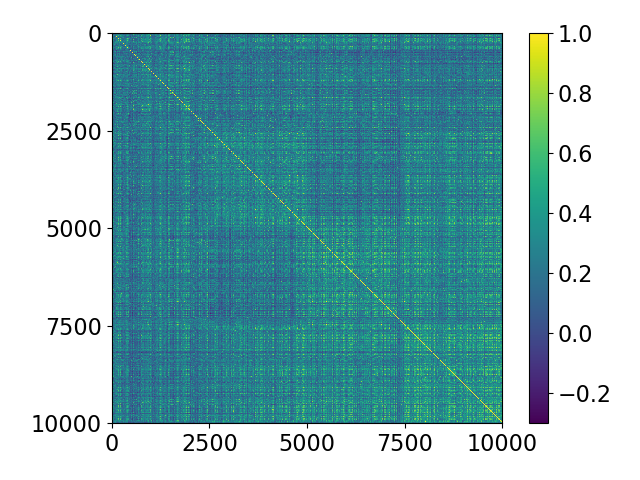}
  \subcaption{Real Data}
  \label{fig:sim1}
\end{subfigure}\hfil
\begin{subfigure}{0.5\textwidth}
  \includegraphics[width=\linewidth]{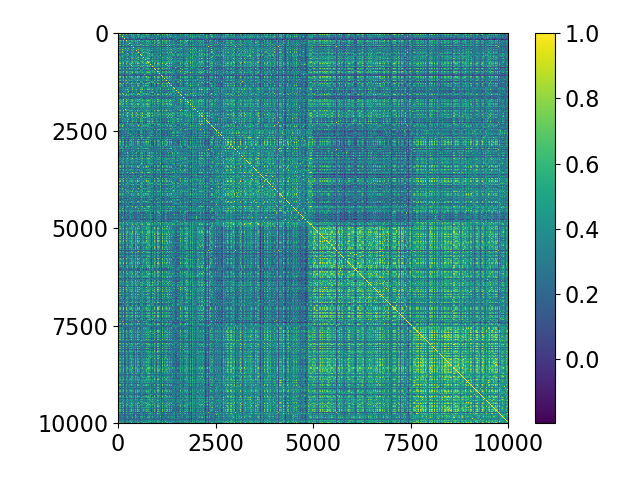}
  \subcaption{Generated Data}
  \label{fig:sim3}
\end{subfigure}
\caption{Similarity matrices for real data and data generated by our CGA model controlling the \textsc{sentiment} and \textsc{verb tense} attributes.}\label{fig:similarity_matr2}
\end{figure*}

\begin{figure*}[h!]
    \centering 
\begin{subfigure}{0.5\textwidth}
  \includegraphics[width=\linewidth]{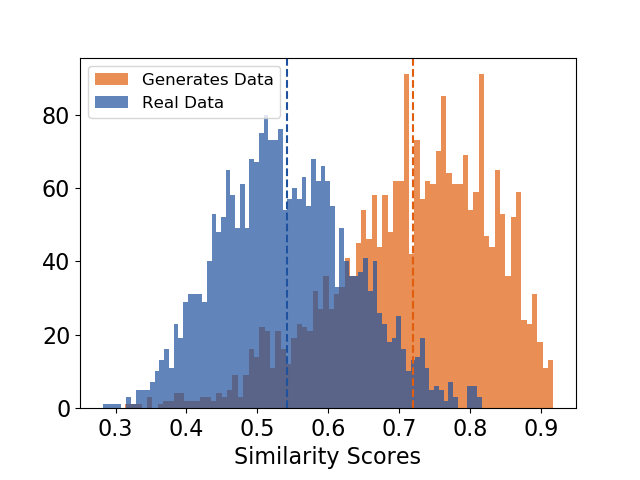}
  \caption{Negative \& Present}
  \label{fig:sim1}
\end{subfigure}\hfil 
\begin{subfigure}{0.5\textwidth}
  \includegraphics[width=\linewidth]{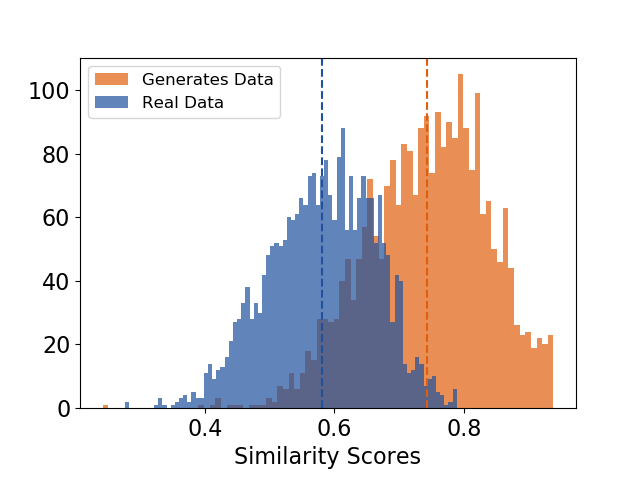}
  \caption{Negative \& Past}
  \label{fig:sim2}
\end{subfigure}\hfil 
\begin{subfigure}{0.5\textwidth}
  \includegraphics[width=\linewidth]{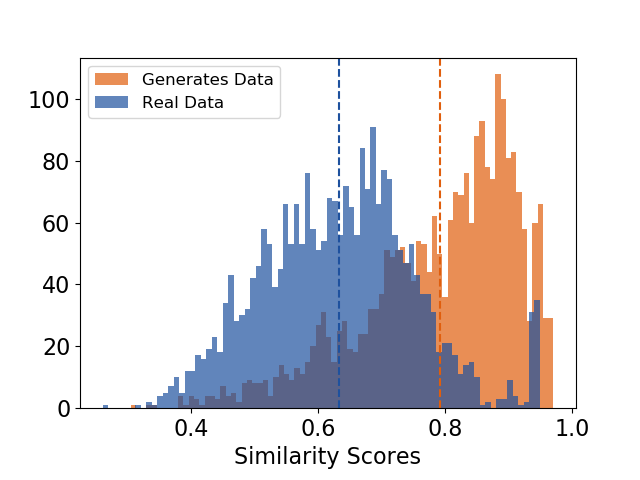}
  \caption{Positive \& Present}
  \label{fig:sim3}
\end{subfigure}\hfil
\begin{subfigure}{0.5\textwidth}
  \includegraphics[width=\linewidth]{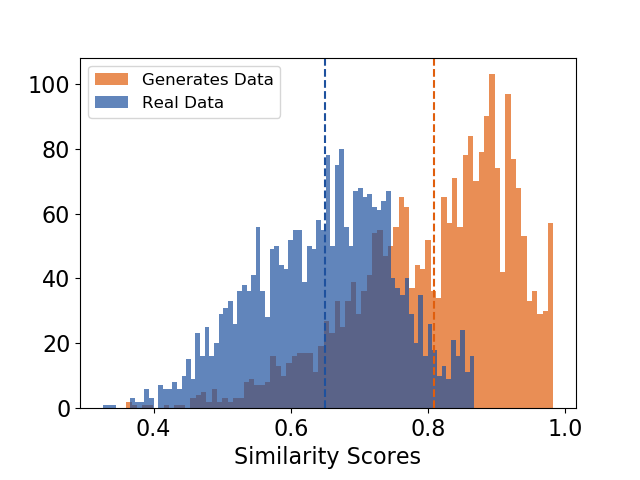}
  \caption{Positive \& Past}
  \label{fig:sim3}
\end{subfigure}
\caption{Sentence similarity scores computed for real data and data generated by our CGA model on the \textsc{sentiment} and \textsc{verb tense} clusters.}\label{fig:sim-histo2}
\end{figure*}

\begin{figure*}[ht!]
    \centering 
\begin{subfigure}{0.5\textwidth}
  \includegraphics[width=\linewidth]{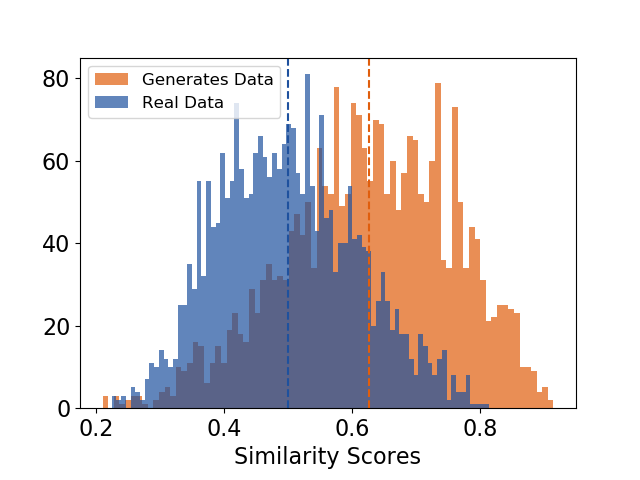}
  \caption{Negative\&Present-Positive\&Present}
  \label{fig:sim1}
\end{subfigure}\hfil 
\begin{subfigure}{0.5\textwidth}
  \includegraphics[width=\linewidth]{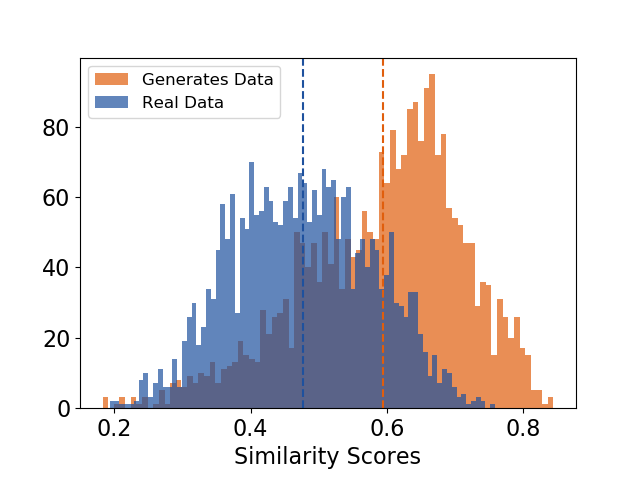}
  \caption{Negative\&Present-Positive\&Past}
  \label{fig:sim2}
\end{subfigure}\hfil 
\begin{subfigure}{0.5\textwidth}
  \includegraphics[width=\linewidth]{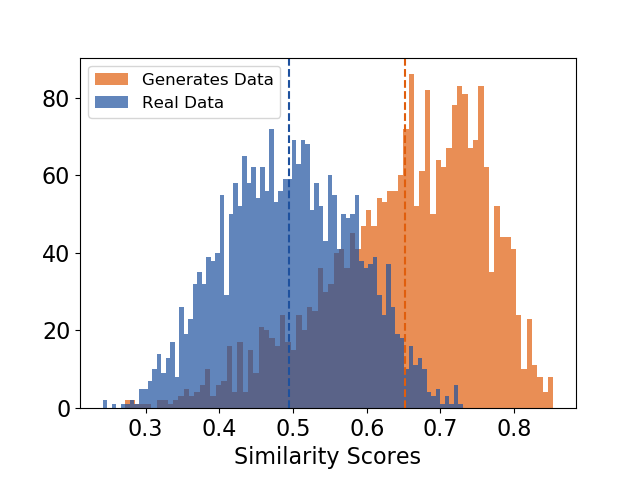}
  \caption{Negative\&Present-Negative\&Past}
  \label{fig:sim3}
\end{subfigure}\hfil
\begin{subfigure}{0.5\textwidth}
  \includegraphics[width=\linewidth]{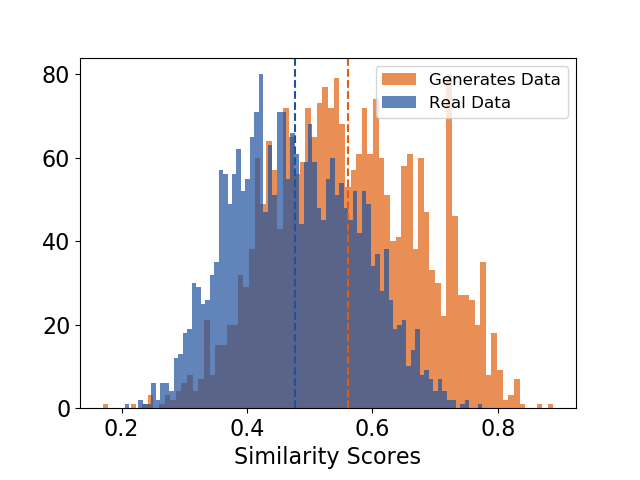}
  \caption{Negative\&Past-Positive\&Present}
  \label{fig:sim3}
\end{subfigure}\hfil
\begin{subfigure}{0.5\textwidth}
  \includegraphics[width=\linewidth]{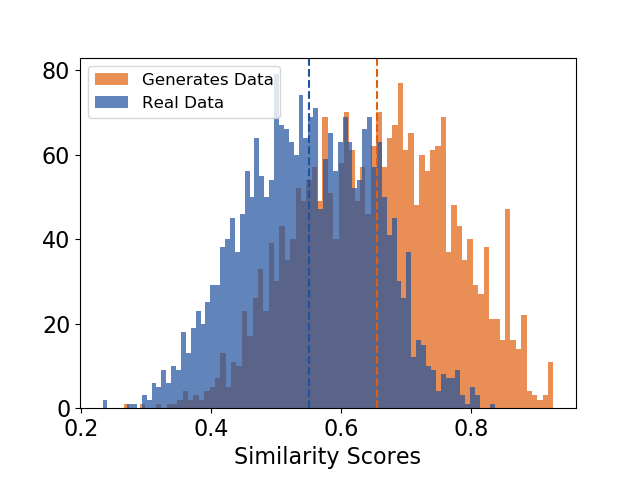}
  \caption{Negative\&Past-Positive\&Past}
  \label{fig:sim3}
\end{subfigure}\hfil
\begin{subfigure}{0.5\textwidth}
  \includegraphics[width=\linewidth]{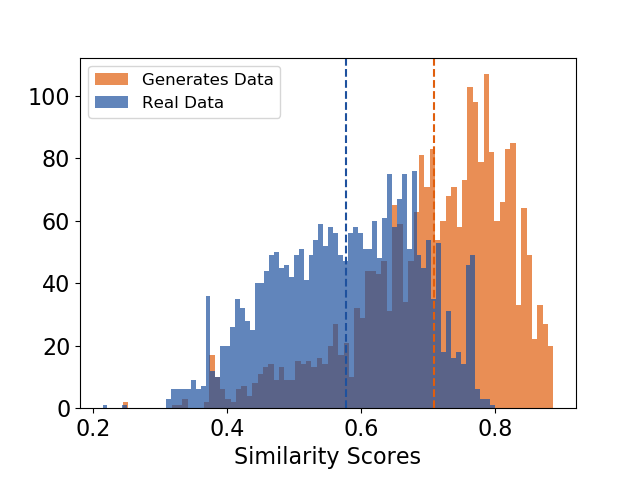}
  \caption{Positive\&Present-Positive\&Past}
  \label{fig:sim3}
\end{subfigure}\hfil
\caption{Sentence similarity scores computed for real data and data generated by our CGA model on the six multi-attribute clusters.}\label{fig:sim-histo3}
\end{figure*}

\subsection{Training Details}\label{suppl:training}

All hyper-parameters were manually tuned. We report the tested ranges in square brackets.

\paragraph{VAE architecture}
Our VAE has one GRU encoder and one GRU decoder.
The encoder has a hidden layer of 256 dimensions [64, 128, 256, 512], linearly transferred to the content vector of 32 dimensions (for one or two attributes), or 50 dimensions (for three attributes) [32, 50, 64, 128].
For training the decoder we set the initial hidden state as $h = Linear(z \oplus a)$. Moreover, we use teacher-forcing combined with the cyclical word-dropout described in Equation \ref{eq:wd}. 

\paragraph{Discriminator}
The discriminator is used to create the attribute-invariant content vectors. We experimented with two architectures for the discriminator which held similar results. We tried a two-layer (64 dimensions each) fully-connected architecture with batch normalization; and a single-layer LSTM with 50 dimensions (for one or two attributes), or 64 dimensions (for three attributes).

\paragraph{KL-Annealing} One of the challenges during the training process was the posterior collapse of the KL term. Similar to \newcite{bowman2016generating}, we used a logistic KL annealing:

\begin{equation}
    \begin{aligned}
        \lambda_{kl}= \frac{1}{1 + exp(-K(x-x_{0}))}
    \end{aligned}{}
\label{eq:kl}
\end{equation}{}

\noindent where $x$ is the current training step. $x_{0}$ indicates how many training steps are needed to set $\lambda_{kl}=1$. $K$ is a constant value given by:

\begin{equation}
    \begin{aligned}
        K = -\frac{log(-1+\frac{1}{1-\epsilon})}{0.5*steps}
    \end{aligned}{}
\label{eq:k}
\end{equation}{}

We set $x_{0}=1000$ for YELP and $x_{0}=5000$ for IMDB. $\epsilon$ is a constant we set to $10^{-4}$.

\paragraph{Discriminator Weight}

The interaction between the VAE and the Discriminator is a crucial factor for our model. Thus,  we decide to linearly increase the discriminator weight $\lambda_{disc}$ during the training process:

\begin{equation}
\small
\begin{aligned}
        \lambda_{disc}(x) =
        \left\{ \begin{array}{ll}
            0 & x\leq k_{1} \\
            min(t, (t / (x_{0})) * (x-k_{1})) & otherwise
        \end{array} \right.
\label{eq:wd2}
\end{aligned}
\end{equation}

 where $t$ is the maximum value that $\lambda_{disc}$ can have.  $x_{0}$ indicates after how many training steps $\lambda_{disc} = t$. $x$ is the current training step. $k_{1}$ is the warm-up value and it indicates after how many training steps the $L_{disc}$ is included in $L_{CGA}$. We set $t=20$, $x_0=6K$  and $k_{1}=12K$ for YELP or $x_{0}=3K$ and $k_{1}=5K$.

\paragraph{Word-Dropout} We use Equation \ref{eq:wd} with the following parameters: $\tau=500$, $k_{min}=0.3$, $k_{max}=0.7$ and \textit{warm-up threshold = 2000}.

\paragraph{Optimizer}
The Adam optimizer, with initial learning rates of $10^{-3}$ [$10^{-5}$, $10^{-4}$, $10^{-3}$, $10^{-2}$, $10^{-1}$], was used for both the VAE and the discriminator \cite{kingma2014adam}.

\subsection{Evaluation}

\subsubsection{Sentence Embedding Similarity}\label{suppl:embed}

Following the approach described in the main paper, we report the results of the sentence embedding similarities for the multi-attribute case (\textsc{sentiment} and \textsc{verb tense}). 
Similarly to the similarity matrices for the single-attribute case, in Figure \ref{fig:similarity_matr2} we recognize the clustered structure of the similarities. These matrices can be divide into the following clusters:

\begin{itemize}
    \item \textbf{Intra-class Clusters:} These are the clusters which are placed over the diagonal of the matrices and show a high cosine similarity scores. They contain similarity scores between the embeddings of samples with the same labels.
    \item \textbf{Cross-Class Clusters:} These are the clusters located above the intra-class clusters. They contains the similarity scores between embeddings of samples with different labels. Indeed, they show lower similarity scores.
\end{itemize}{}

To gain a qualitative understanding of the generation capacities of our model, we assume that an ideal generative model should produce samples that have comparable similarity scores to the ones of the real data. We contrast the similarity scores computed on each cluster separately in the histograms in Figures \ref{fig:sim-histo2} and \ref{fig:sim-histo3}.

\begin{table*}[ht]
\centering
\begin{tabular}{|l|cccccccccc|}
\hline
\textbf{Augmentation} (\%) & \textbf{0} & \textbf{10} & \textbf{20} & \textbf{30} & \textbf{50} & \textbf{70} & \textbf{100} & \textbf{120} & \textbf{150} & \textbf{200} \\\hline\hline
\textbf{500 sentences} &  &  &  &  &  &  &  &  &  &  \\
Real Data & 0.753 & 0.759 & 0.770 & 0.772 & 0.781 & 0.792 & 0.797 & 0.797 & 0.806 & 0.814 \\
Generated Data & - & 0.772 & 0.773 & 0.772 & 0.781 & 0.793 & 0.785 & 0.795 & 0.802 & 0.803 \\
EDA & - & 0.753 & 0.755 & 0.759 & 0.762 & 0.762 & 0.732 & 0.732 & 0.730 & 0.749 \\\hline
\textbf{1000 sentences} &  &  &  &  &  &  &  &  &  &  \\
Real Data & 0.794 & 0.791 & 0.801 & 0.798 & 0.818 & 0.813 & 0.829 & 0.820 & 0.834 & 0.847 \\
Generated Data & - & 0.806 & 0.812 & 0.811 & 0.812 & 0.811 & 0.815 & 0.821 & 0.822 & 0.819 \\
EDA & - & 0.787 & 0.785 & 0.792 & 0.780 & 0.780 & 0.767 & 0.771 & 0.760 & 0.756 \\\hline
\textbf{10000 sentences} &  &  &  &  &  &  &  &  &  &  \\
Real Data & 0.877 & 0.886 & 0.883 & 0.883 & 0.884 & 0.881 & 0.890 & 0.894 & 0.896 & 0.897 \\
Generated Data & - & 0.874 & 0.878 & 0.878 & 0.873 & 0.875 & 0.878 & 0.877 & 0.877 & 0.878 \\
EDA & - & 0.880 & 0.884 & 0.881 & 0.881 & 0.881 & 0.862 & 0.863 & 0.866 & 0.864 \\\hline
\end{tabular}
\caption{Detailed accuracy numbers for YELP data augmentation results presented in Figure \ref{fig:dataug-yelp-line}.}
\label{tab:yelp-table}
\end{table*}

\begin{table*}[ht]
\centering
\begin{tabular}{|l|cccccccccc|}
\hline
\textbf{Augmentation} (\%) & \textbf{0} & \textbf{10} & \textbf{20} & \textbf{30} & \textbf{50} & \textbf{70} & \textbf{100} & \textbf{120} & \textbf{150} & \textbf{200} \\\hline\hline
\textbf{500 sentences} &  &  &  &  &  &  &  &  &  &  \\
Real Data & 0.548 & 0.555 & 0.535 & 0.558 & 0.559 & 0.574 & 0.571 & 0.568 & 0.576 & 0.569 \\
Generated Data &- & 0.552 & 0.556 & 0.561 & 0.559 & 0.583 & 0.595 & 0.583 & 0.589 & 0.587 \\
EDA & - & 0.537 & 0.551 & 0.548 & 0.534 & 0.534 & 0.546 & 0.550 & 0.562 & 0.540 \\\hline
\textbf{1000 sentences} &  &  &  &  &  &  &  &  &  &  \\
Real Data & 0.570 & 0.573 & 0.574 & 0.561 & 0.591 & 0.577 & 0.591 & 0.598 & 0.598 & 0.615 \\
Generated Data & - & 0.569 & 0.568 & 0.589 & 0.579 & 0.586 & 0.592 & 0.594 & 0.589 & 0.608 \\
EDA & - & 0.570 & 0.582 & 0.555 & 0.561 & 0.561 & 0.568 & 0.544 & 0.559 & 0.558 \\\hline
\textbf{10000 sentences} &  &  &  &  &  &  &  &  &  &  \\
Real Data & 0.663 & 0.674 & 0.672 & 0.672 & 0.670 & 0.672 & 0.675 & 0.676 & 0.673 & 0.676 \\
Generated Data & - & 0.666 & 0.663 & 0.663 & 0.667 & 0.662 & 0.656 & 0.668 & 0.666 & 0.669 \\
EDA & - & 0.664 & 0.662 & 0.662 & 0.672 & 0.672 & 0.652 & 0.658 & 0.653 & 0.651 \\\hline
\end{tabular}
\caption{Detailed accuracy numbers for IMDB data augmentation results presented in Figure \ref{fig:dataug-imdb-line}.}
\label{tab:imdb-table}
\end{table*}

\subsubsection{Data Augmentation}\label{suppl:results}

For the data augmentation experiments we use a bidirectional LSTM with input size 300 and hidden size 256 [64, 128, 256, 512]. The input size is given by the 300 dimensional pre-trained GloVe embeddings \cite{pennington2014glove}. We set dropout to 0.8 [0.5, 0.6, 0.7, 0.8]. For the training we use early stopping, specifically we stop the training process after 8 epochs without improving the validation loss.

Tables \ref{tab:yelp-table} and \ref{tab:imdb-table} show the detailed results for the data augmentation experiments on IMDB and YELP, respectively. The standard deviation of all results over 5 random seeds was always \textless 0.025.

Table \ref{tab:val} shows the corresponding performance on the validation set for the results on the test set presented in Table \ref{tab:dataug-best}.

\begin{table*}[ht]
\centering
\begin{tabular}{|l|c|c|l|c|l|c|}
\hline 
                  & \multicolumn{6}{c|}{\textbf{Training Size}}                                                            \\
\textbf{}         & \multicolumn{2}{c|}{\textbf{500 sentences}} & \multicolumn{2}{l|}{\textbf{1000 sentences}} & \multicolumn{2}{l|}{\textbf{10000 sentences}} \\ \hline\hline
\textbf{Model}    & acc. (std)                          & \%    & \multicolumn{1}{c|}{acc. (std)}     & \%     & \multicolumn{1}{c|}{acc. (std)}      & \%     \\ \hline
Real Data YELP    & 0.79 (0.05)                         & 0     & \multicolumn{1}{c|}{0.77 (0.04)}    & 0      & \multicolumn{1}{c|}{0.86 (0.04)}     & 0      \\ 
YELP + EDA        & 0.77 (0.09)                         & 70    & \multicolumn{1}{c|}{0.82 (0.04)}    & 70     & \multicolumn{1}{c|}{\textbf{0.87} (0.03)}     & 70     \\ 
YELP + CGA (Ours) & \textbf{0.81} (0.03)                         & 150   & \multicolumn{1}{c|}{\textbf{0.84} (0.03)}    & 120    & \multicolumn{1}{c|}{0.85 (0.07)}     & 100    \\ \hline\hline
Real Data IMDB    & \multicolumn{1}{l|}{0.55 (0.03)}    & 0     & 0.57 (0.04)                         & 0      & 0.65 (0.09)                          & 0      \\ 
IMDB + EDA        & \multicolumn{1}{l|}{0.56 (0.04)}    & 150   & 0.58 (0.06)                         & 70     & \textbf{0.69} (0.05)                          & 100    \\ 
IMDB + CGA (Ours) & \multicolumn{1}{l|}{\textbf{0.62} (0.05)}    & 120   & \textbf{0.59} (0.03)                         & 200    & 0.68 (0.02)                          & 120    \\ \hline
\end{tabular}
\caption{Performance on the validation set for the data augmentation results reported in Table \ref{tab:dataug-best}.}\label{tab:val}
\end{table*}

\subsubsection{TextCNN} 
For the Attribute Matching results presented in Section \ref{sec:eval} we use the pre-trained TextCNN \cite{kim2014convolutional}. This network standardly uses 100 dimensional Glove word embeddings \cite{pennington2014glove}, 3 convolutional layers with 100 filters each. The dropout rate is set to 0.5 during the training process.

\subsection{Generated Sentences}\label{suppl:sents}
Tables \ref{tab:examples-sentiment} to \ref{tab:examples-person} provide example sentences generated by the CGA model for the three individual attributes. Moreover, the code repository provides all generated sentences\footnote{Link omitted for review}.

\subsection{Computing Infrastructure}
All models presented in this work were implemented in PyTorch, and trained and tested on single Titan XP GPUs with 12GB memory.

The average runtime was 07:26:14 for the model trained on YELP. The average runtime was 04:09:54 for the model trained on IMDB.

\begin{table*}[ht!]
\centering
\begin{tabular}{ll}
\textbf{Sentence} & \textbf{Sentiment} \\\hline
but i'm very impressed with the food and the service is great. & Positive \\
i love this place for the best sushi! & Positive \\
it is a great place to get a quick bite and a great price. & Positive \\
it's fresh and the food was good and reasonably priced. & Positive \\
not even a good deal. & Negative \\
so i ordered the chicken and it was very disappointing. & Negative \\
by far the worst hotel i have ever had in the life. & Negative \\
the staff was very rude and unorganized. & Negative \\\hline
\end{tabular}
\caption{Examples of generated sentences controlling the \textsc{sentiment} attribute.}
\label{tab:examples-sentiment}
\end{table*}

\begin{table*}[t!]
\centering
\begin{tabular}{ll}
\textbf{Sentence} & \textbf{Tense}  \\\hline
i love the fact that they have a great selection of wines. & Present \\
they also have the best desserts ever. & Present \\
the food is good , but it's not worth the wait for it. & Present \\
management is rude and doesn't care about their patients. & Present \\
my family and i had a great time. & Past \\
when i walked in the door , i was robbed. & Past \\
had the best burger i've ever had. & Past \\
my husband and i enjoyed the food. & Past \\\hline
\end{tabular}
\caption{Examples of generated sentences controlling the \textsc{verb tense} attribute.}
\label{tab:examples-tense}
\end{table*}

\begin{table*}[t!]
\centering
\begin{tabular}{ll}
\textbf{Sentence} & \textbf{Person} \\\hline
it was a little pricey but i ordered the chicken teriyaki. & Singular \\
she was a great stylist and she was a sweetheart. & Singular \\
worst customer service i've ever been to. & Singular \\
this is a nice guy who cares about the customer service. & Singular \\
they were very friendly and eager to help. & Plural \\
these guys are awesome! & Plural \\
the people working there were so friendly and we were very nice. & Plural \\
we stayed here for \textsc{NUM} nights and we will definitely be back. & Plural \\\hline
\end{tabular}
\caption{Examples of generated sentences controlling the \textsc{person number} attribute.}
\label{tab:examples-person}
\end{table*}

\end{document}